\documentclass[a4paper,man,natbib,floatsintext]{apa7}
\usepackage{graphicx} 

\usepackage{amsmath}
\usepackage{amssymb}
\usepackage{graphicx}

\usepackage{pslatex}
\usepackage{apacite}
\usepackage{float} 

\title{Analogy as Nonparametric Bayesian Inference over Relational Systems}
\shorttitle{Analogy as nonparametric Bayesian inference}

\authorsnames[1,2]{Ruairidh M. Battleday, Thomas L. Griffiths}

\authorsaffiliations{{Department of Psychology, Harvard University}, {Department of Psychology, Princeton University}}
\abstract{Our inferences in the real world are rarely na{\"i}ve---we acquire experiences through our lifetime that can help us more quickly understand the structure of something new. A fundamental question in cognitive science is how we make such generalizations. Studies of analogy have explored the question of how to map information from a single familiar concept or environment to an unfamiliar one. In this paper, we examine how experience with multiple successive environments affects an individual's subsequent inferences. First, we present an online behavioral environment in which participants play a number of virtual games that each operate according to an underlying relational structure. Second, we show that exposing participants to a particular relational structure biases them towards expecting the same structure to hold in the test game, an effect that scales with the number of times the structure has been observed. Finally, we propose a novel probabilistic model that accounts for these behaviors in terms of nonparametric Bayesian inference. This model generates predictions from each past environment based on their relational structures, and then averages  predictions from individual environments according to their analogical relevance to the task at hand. Our results and statistical framework provide a complementary perspective for several key computational ideas about analogy, and our nonparametric framework allows us to account for how a learner might continually build and use knowledge over a lifetime.}


\begin{document}
\maketitle

\section{Introduction}
Consider a hunter-gatherer foraging in a newly discovered valley. The plants and animals in this ecosystem are likely to interact in particular ways. How can the hunter-gatherer identify which interactions to target, in order to ensure the best chance of a feed? Or, which should not be disrupted, to maintain a healthy ecosystem or avoid predation? 

Now consider a recent graduate trying to decide whom to approach about a problem on their first day of work. It seems likely that different people within the office will be more or less helpful for different types of problem. They have seen and heard several of their new colleagues interacting: which colleague is most likely to be helpful; and for which problems? 

These two scenarios are cases of \textit{relational generalization}: the computational problem of deciding which unobserved interactions within an environment are likely to occur based on the interactions observed so far. Although the \textit{features} of both situations are important, using them exclusively as the basis for generalization would provide a poor model of their underlying consequential structures. Many plants produce brightly colored flowers; but it is their relationship to the other elements of the environment that helps predict whether they are poisonous or not---their physical relation to other plants and animal tracks; and to the climate, seasons, and weather. Equally, an executive manager and a receptionist may smile the most, and have the most open office plans. But neither should be approached for specific technical questions. Instead, the categories of the workplace are better organized according to \textit{relational} principles: Who interacts with whom; and with what frequency? About what; and where do different sets of expertise intersect?

Without background knowledge, solving these kinds of problems would be hard---we would have to observe enough to begin to form a theory of how the new environment works; or be lucky with our na{\"i}ve guesses. Fortunately, in the real world we often have related knowledge that we can \textit{generalize} from other environments. This includes the two scenarios listed at the start, where the gatherer will have relevant prior knowledge about foodchains and ecology abstracted from many different valleys; and the graduate about hierarchies and the relationship between responsibilities and skill sets abstracted from classrooms and weekend jobs.

This paper outlines a computational approach to studying such behavior. We present an experimental paradigm in which participants move colored blocks and explore how they interact. Participants can experience an arbitrary number of these environments, with arbitrary features and underlying dynamics. We find that participants perform better in a new environment if they have been previously exposed to its relational structure. This replicates a seminal result on analogy discovered in the verbal domain, where participants told a background story are more likely to solve an analogous problem in a different story setting if the stories are structurally similar \citep{gick1980analogical}. We go on to show that this effect \textit{increases} with the number of times a relational structure has been encountered by a participant in the past. 

Our key theoretical contribution is to account for this generalization behavior using nonparametric Bayesian inference. We capture the relational structures of different training environments using mathematical abstractions of intuitive theories \citep{carey1986cognitive}; and in particular stochastic blockmodels \citep{holland1983stochastic}. Predictions can be made at test time by Bayesian model averaging over these abstractions, such that individual predictions from different environments are combined according to their \textit{relevance}---or, their posterior probability given the data from the new environment. 

Our framework has many close connections to structure-mapping theory (SMT), an influential computational account of analogical transfer \citep{gentner1983structure}. In SMT, the goal of analogy is to find a \textit{mapping} between two knowledge structures, such that inferences about missing data in the target structure can be made from analogous relations in the base. SMT posits that a number of computational constraints help a learner make such mappings. One such requirement is that the mapping be \textit{one-to-one}: the classes and relations in the base environment must each be mapped to a unique class or relation in the target. Our framework assigns \textit{elements} in the new system to the \textit{classes} of previously encountered relational structures, allowing many-one, one-one, and one-many mappings, but weighting them by their ability to explain the data observed so far. Another important theoretical contribution of SMT is the idea of ``systematicity,'' which upweights mappings that connect more deeply embedded structures. Our identification of analogical relevance with the posterior probability of a knowledge representation corroborates this principle---a match between two systems will be more systematic and receive higher posterior probability if their generative structures run deeper \citep{kemp2005generative}. Our approach builds on SMT in showing how to integrate information from \textit{multiple} different concepts or environments into a learner's generalizations.

In the next section, we ground the idea of information transfer between different learning environments in the formal framework of generalization. This allows us to both propose computational models for human inference in non-n{\"a}ive situations, including the examples given above, as well as test and extend aspects of models on analogy. We then present a behavioral paradigm that can be used to test how experience with the structures of different environments influences participant's predictions. Our main hypothesis is that an increasing number of exposures to a particular relational structure during training will increasingly bias participants to expect that same structure in a test environment. We find that this hypothesis holds in a large-scale online experiment, and show that its results can be captured by our nonparametric Bayesian framework. 

\section{The computational problem of relational generalization}
The ecological importance of generalization as a computational problem was first highlighted by Shepard (\citeyear{shepard1987toward}), who argued that because no two situations could ever be alike a satisfying solution to the problem of generalization would constitute psychology's first universal law. Shepard showed that for a wide range of stimuli participants' generalizations were remarkably consistent, and could be related to how \textit{similar} pairs of stimuli were---or, how conceptually close they were. For these stimuli, the probability of generalizing a consequential property decreased according to the negative exponential of the distance between the stimuli in psychological space. Differences between individuals could then be explained by having different psychometric functions for embedding stimuli into this space. 

Tenenbaum and Griffiths (\citeyear{tenenbaum2001generalization}) extended Shepard's account to consider how the generalization of consequential properties changed with exposure to multiple objects possessing it, or if the properties identifying objects were discrete instead of continuous. Most importantly, their unification expanded Shepard's characterization of generalization as a problem of probabilistic inference, in which a learner can use abstract representations of the statistical structure underlying an environment to predict whether the property in question will generalize. This more general framing motivated the development of models based on the generalization of \textit{relations}, rather than just continuous or discrete features \citep{kemp2008discovery, kemp2010probabilistic, tenenbaum2006theory}. Relational information is abundant and consequential in all human experience---from social structures, to ecological knowledge, to music and dictionaries \citep{hofstadter2008metamagical}. 


\subsection{Modeling relational generalization}
At an abstract level, we are interested in modeling the probability of an unobserved interaction, $r$, given a set of observed interactions, $\mathcal{R}$:
\begin{equation}
    p(r|\mathcal{R}).
\end{equation}

\noindent This is a conditional probability distribution, and one could provide a model for it in many ways. 

One immediate observation is that the observed interactions may contain patterns that help identify the overall structure of the domain. There is converging evidence from developmental psychology, computational cognitive science, and reinforcement learning that people organize such patterns into ``intuitive theories,'' which help them make sense of novel empirical environments \citep{carey1986cognitive, gopnik1997words, kemp2010probabilistic, tenenbaum2006theory, pouncy2022inductive, tsividis2021human}. These theories contain information about the classes of element in a domain and the ways in which they interact. Consider a child exploring the interactions of certain toys in their room. Such an environment can be succinctly described by the latent classes of objects (for example, ``magnets,'' ``metal objects,'' ``plastic objects''), and the different ways in which they interact (for example, ``repels'' and ``attracts'').

Individual theories can be expressed by a mathematical form known as a stochastic blockmodel, which accounts for interactions between \textit{elements} in terms of how strongly their underlying \textit{classes} relate \citep{holland1983stochastic}. The model is then written as: 
\begin{equation}
    \mathcal{T} \, = \, \{\eta, {\bf z} \},
\end{equation}

\begin{figure}[h!]
    \centering 
\includegraphics[width=0.5\linewidth]{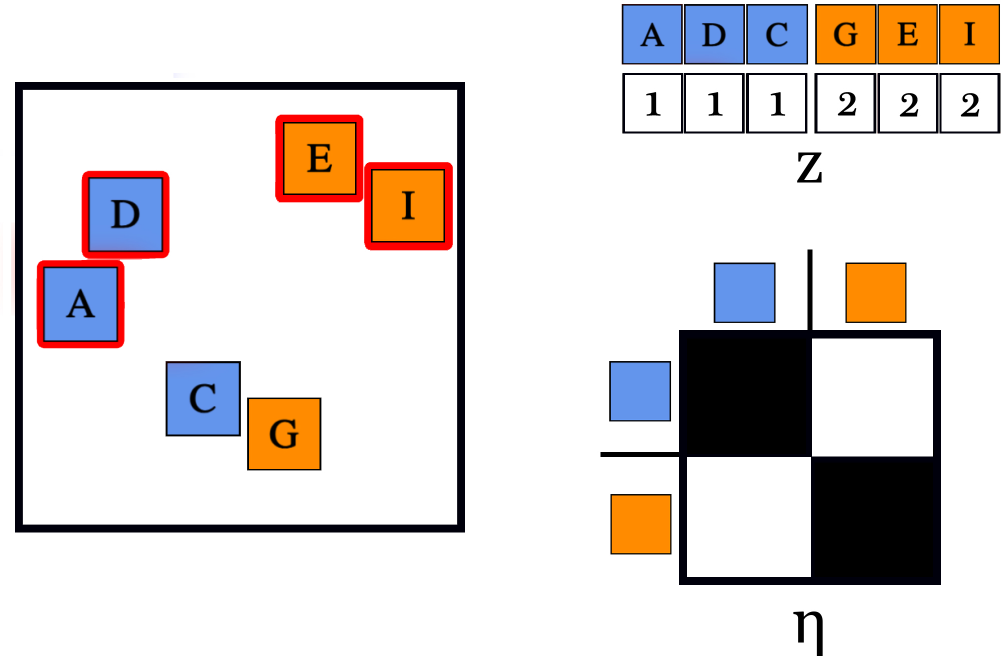}
    \caption{An illustration of the stochastic blockmodel. Left panel: An environment in which pairs of named colored blocks may or may not interact (by turning red when placed next to each other). Right panels: The two components of a stochastic blockmodel, in this case expressing the relational structure of the environment in the left panel. A vector, ${\bf z}$, assigns the named elements to one of two latent classes, in this case determined by their color.  A class-level matrix, $\eta$, then gives the probability of different classes interacting. The entries of the matrix represent whether squares from pairs of color classes interact. If the color classes indicated by the row and column of the matrix always interact, the entry between them will be black (representing an interaction probability of 1). If they never interact, it will be white (representing an interaction probability of 0).}
    \label{fig:blockmodel}
\end{figure}

\noindent Mathematically, it uses a class assignment vector, ${\bf z}$, to allocate each element to a latent class, where the number of latent classes is known in advance. For example, for element $i$, the $i$th entry of ${\bf z}$ will give its latent class ($1$, perhaps). The probability of each pair of classes interacting is modeled explicitly in a matrix containing class-level interaction probabilities, $\eta$. For example, the probability that elements from class $1$ will interact with elements from class $2$ is given by the corresponding matrix entry, $\eta_{1, 2}$. These two components constitute the mathematical representation of a stochastic blockmodel; or, an individual intuitive theory, $\mathcal{T}$ (see Figure \ref{fig:blockmodel}). Allocations and class-level probabilities can be input into the model, or estimated based on the proportion of observed interactions.

The Infinite Relational Model (IRM; \citeauthor{kemp2010probabilistic}, \citeyear{kemp2010probabilistic}) extends stochastic blockmodels to express intuitive theories that account for how an adaptive learner should generalize when exploring a new environment. The learner begins with a set of prior beliefs, encoded as a probability distribution, over every possible theory (each expressed in the form of a stochastic blockmodel). It then updates these beliefs in light of the observed interactions according to Bayes' rule. The learner can then make predictions over unobserved interactions by averaging the predictions of all possible theories by their posterior probability. 

With this notation in hand, the IRM provides a starting point to compute the conditional probability of any particular generalization:
\begin{equation}
   p(r|\mathcal{R}) =  \int_{\eta'} \sum_{{\bf z}'} p(r| {\bf z}', \eta')\cdot p({\bf z}', \eta'|\mathcal{R})  \, d\eta'.
\end{equation}

\noindent This equation states that a prediction about the unobserved interaction can be decomposed into a prediction made under each individual theory that is then weighted by the posterior probability of that theory given the observed data. Integrating across all possible theories will give the final conditional probability we require---a process more generally known as\textit{ hypothesis averaging}, where here the generative models provided by individual theories or blockmodels are our hypotheses for the data. The unobserved relation(s), $r$, are now considered independent from the observed relations, $\mathcal{R}$, given the model parameters, ${\bf z}, \, \eta$. As the IRM maintains a distribution over every possible theory it can grow indefinitely in complexity with the number of observed elements. Since it is not constrained to any parametric class of models, the resulting model is \textit{nonparametric}; in this case we will refer to the IRM as being nonparametric in the number of elements, as more elements result in more complexity. 

\section{Generalizing between environments}
So far, we have considred how relations can be generalized within a single environment. The majority of real-world generalization, however, occurs in partially explored environments where a learner has had some past experience that is related to the environment or task at hand. This kind of background information becomes increasingly important as the number of observations decreases, or if the structure underlying an environment learner is less typical or more complex. In this section, we show how to extend the formalism given by intuitive theories and the IRM to capture experience derived from past learning environments. An optimal learner should be able to combine this information with the observations provided by the new environment to make the best inferences possible; and then update their knowledge as new information accrues. 

\subsection{Modeling past environments}
Our strategy is to provide the learner with a set of ``models,'' each representing knowledge they have abstracted from a past environment: $\mathcal{M} \equiv \{M\}_{k=1}^K$. A generalization is the result of marginalizing over these models:
\begin{equation}
    p(r|\mathcal{R}) = \sum_{M' \in \mathcal{M}} p(r|\mathcal{R}, M') \cdot p(M'|\mathcal{R}).
\end{equation}

\noindent This defines a mixture distribution, where the kernel---the conditional probability given a model, $p(r|\mathcal{R}, M)$---is weighted by the mixture weight or model posterior, $p(M|\mathcal{R})$.

To compute each of these terms, we need to introduce the parameters that bind the models to the data. We represent all of a model's parameters a single parameter, $\theta$:
\begin{equation}
   p(r|\mathcal{R}, M) =  \int_{\theta'} p(r| \theta', M)\cdot p(\theta'|\mathcal{R}, M) \, d\theta'.
\end{equation}
%

\noindent We combine these expansions to arrive at the full expression for the conditional probability of generalization:
\begin{align}
    p(r|\mathcal{R}) & = \sum_{M' \in \mathcal{M}} p(r|\mathcal{R}, M')\cdot p(M'|\mathcal{R})  \\
    &=\sum_{M' \in \mathcal{M}} \int_{\theta'} p(r| \theta')\cdot p(\theta'|\mathcal{R}, M') d \theta' \cdot p(M'|\mathcal{R}).
\end{align}

This expression models a rational learner's generalization as a hierarchical process of probabilistic inference, in which the predictions are first computed from the parameters of individual models (hypothesis averaging), and then combined across all of the learner's different models (model averaging). Individual predictions from a particular model are weighted by how well they capture the observed data. The overall prediction of a particular model is in turn weighted by how relevant it is to the novel environment. The weights in both cases are given by the following posterior distributions, which can be calculated using Bayes' rule:
\begin{align}
    p(M|\mathcal{R}) &= \frac{p(\mathcal{R}|M) \cdot p(M)}{p({\mathcal{R}})};\\
     p(\theta|\mathcal{R}, M) &= \frac{p(\mathcal{R}|\theta) \cdot p(\theta|M)}{p({\mathcal{R}}|M)}.
\end{align}

The term $p(\mathcal{R}|M)$ is known as the Bayes' factor or ``model evidence.'' It can be computed by integrating out the model's parameters, or estimated in various ways \citep{kass1995bayes}. The prior distribution over models, $p(M)$, allows us to incorporate a variety of relevant information---for example, the number of times a model has been encountered in the past, or the probability of an entirely new theory.

\subsection{Nonparametric Bayesian inference over structures}
At a high level, our framework defines a \textit{nonparametric} approach to relational generalization, in which the inferences drawn by a learner in a new environment depend on a distribution over theories arrived at through experience. Unlike the IRM, which is nonparametric over the number of elements in an environment, our framework is nonparametric in the number of structures---or, ``models''---a learner uses to generalize. Using a nonparametric distribution as the basis for inference is a common technique in both frequentist and Bayesian statistics, and has a long history of use in the study of human categorization in the form of exemplar models \citep{nosofsky1986attention}. We then use Bayesian model averaging to collapse this distribution and make generalizations about the new environment. Intuitively, our model assumes people remember past relational structures and use them to predict new relations, weighting each structure by its similarity to the current environment.

The remaining challenge is to specify how these abstract models might each express the information a learner has abstracted from a past environment. In this paper, we consider environments whose interactions are overwhelmingly dominated by a prominent relational structure (consider the magnetic, ferrous, and plastic toys presented above). Formally, we consider an abstraction of a past environment, $M$, to consist of the number of latent classes, the frequency with which they arise, and counts of how often elements from those classes did or did not interact under a particular relation. This information is encoded in a class-level frequency vector, $f$, and class-level count, $u$, and non-count, $\bar{u}$ matrices for each type of relation; $M\equiv\{f, u_1, \bar{u}_1, u_2, \bar{u}_2, \ldots, \}$. This single relational structure can be used to nucleate a prior distribution (or, kernel) over theories about the new environment---in this case, the \textit{a priori} probabilities of all possible assignments of the elements in the new system to the existing kernel's relational structure. The observed data from the new environment can be used to update this to a posterior distribution (see Methods for further details). Averaging over different models based on their posterior probability will upweight predictions based on theories that express salient information shared between the new and old environments. 

\section{Testing the model predictions}
We want to test the idea that prior experience leads to abstractions that affect---help or hinder---inference in a novel environment. This is, in essence, the effect observed in the seminal study of analogy by \citeauthor{gick1980analogical} (\citeyear{gick1980analogical}), who found that participants provided better solutions to a verbal problem if they were first told a story with a similar relational structure. Because it is nonparametric our framework makes an additional prediction: that this effect will increase according to the \textit{number} of times a particular structure has been encountered in different training environments. To test this novel prediction, we need an experimental setup where we can vary the number of prior exposures a participant has to a consequential relational structure. We would also like to be able to vary the visual features of different environments independently from the relational structure, to disambiguate the effects of similar features and similar relations.

\begin{figure}[h!]
    \centering 
    \includegraphics[width=0.6\linewidth]{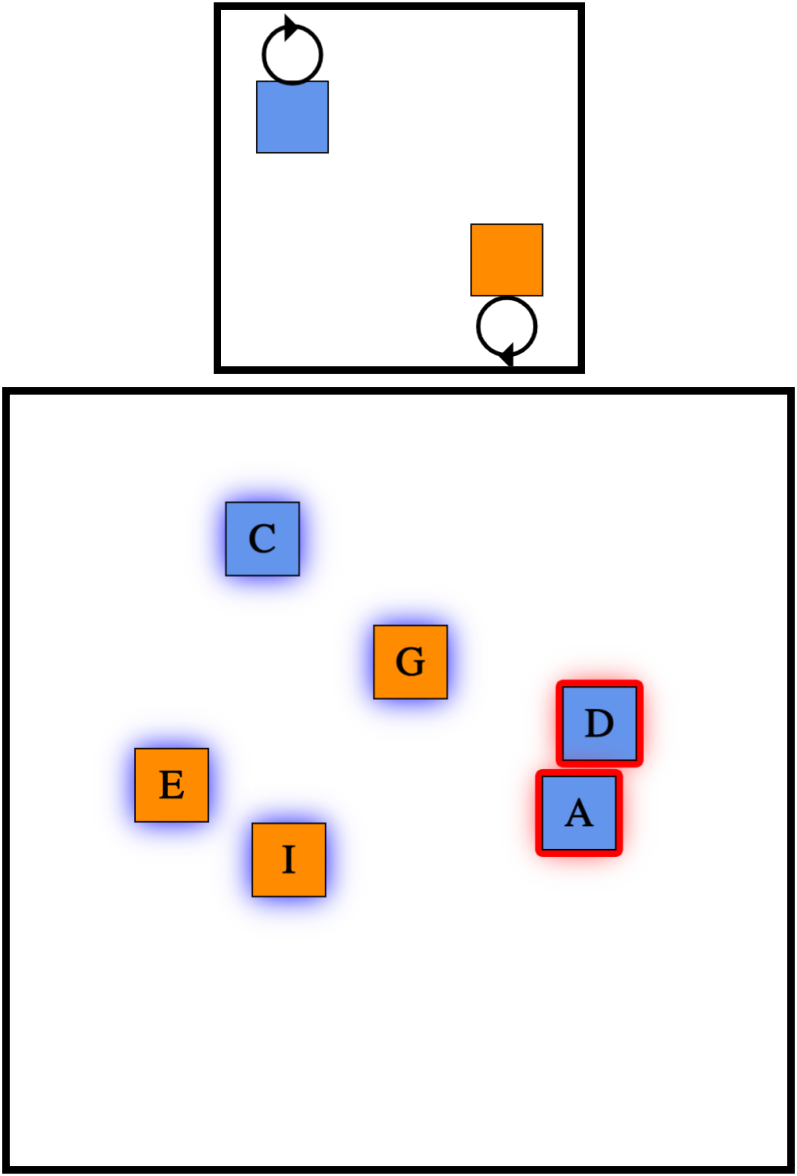}
    \caption{A blockworld. The game environment contains colored, movable squares that may or may not interact when they come into contact. There are two types of trial in each game: the manipulate phase (shown above), in which all objects have a blue shadow and can be moved freely for a untimed period; and the rate phase, in which the objects locations are fixed and their blue shadows are removed, and participants asked questions about whether one of a pair of objects would interact (see Figure \ref{fig:demo_testing_rate}). In ``training'' games (shown above), a set of instructions appear above the main environment, containing schematics of the game structure. In the testing games, these instructions are removed from the display (see Figure \ref{fig:demo_testing_rate}).}
    \label{fig:demo_training}
\end{figure}

For this, we introduce a virtual experimental platform called a \textit{blockworld} (inspired by \citeauthor{kemp2010probabilistic}, \citeyear{kemp2010probabilistic}; see Figure \ref{fig:demo_training}). Participants are told that they will be a worker at a toy factory and will have to test whether a number of games (blockworlds) work properly in order to approve them for production and proceed in the experiment. They will be paid for each game they approve, and receive bonuses according to how well they understand the workings---or, structure---of each game. Our purpose is to show that training on games with different underlying structures will lead to different inferences in a test game. If they share the same structure, we predict that participants will be able to make better inferences earlier. If they are incongruous, inference should not change much from baseline. And if they conflict, performance will decrease.

\subsection{Experimental setup}
\begin{figure}[h!]
    \centering 
    \includegraphics[width=0.6\linewidth]{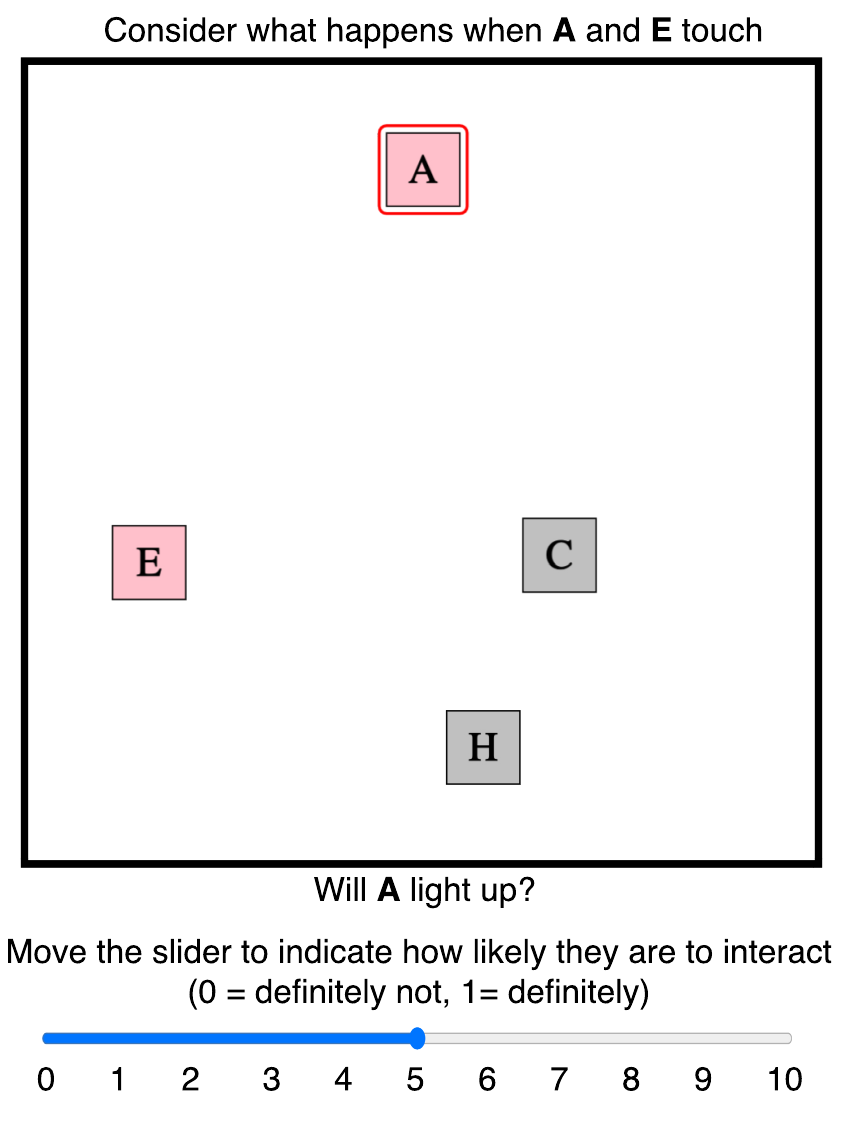}
    \caption{An example rate phase from a testing game (round 1). Here, the question being asked is ``same new'': whether the probe object, ``A,'' will be lit up by the existing block of the same color group, ``E.''}
    \label{fig:demo_testing_rate}
\end{figure}

Each blockworld contains a set of colored labeled squares. When the squares are movable they are outlined with a soft blue shadow. When the user touches two squares together, one or both of them will ``light up,'' meaning their outlines will turn opaque and red. Each game comprises a ``manipulate'' phase and a set of ``rate'' phases, all of which occur within the same blockworld. In the untimed manipulate phase (Figure \ref{fig:demo_training}), participants are instructed to ``play with the blocks and see what lights up.'' When they are ready, they click a button to continue. The subsequent rate phases ask participants to rate whether particular pairs of blocks will light up when they touch (Figure \ref{fig:demo_testing_rate}). Participants indicate their response using a slider, with buttons labeled with increasing integers from 0 (``definitely not interact'') to 10 (``definitely interact''). After the questions, they can freely move the blocks again. 

\begin{figure}[h!]
    \centering 
\includegraphics[width=0.7\linewidth]{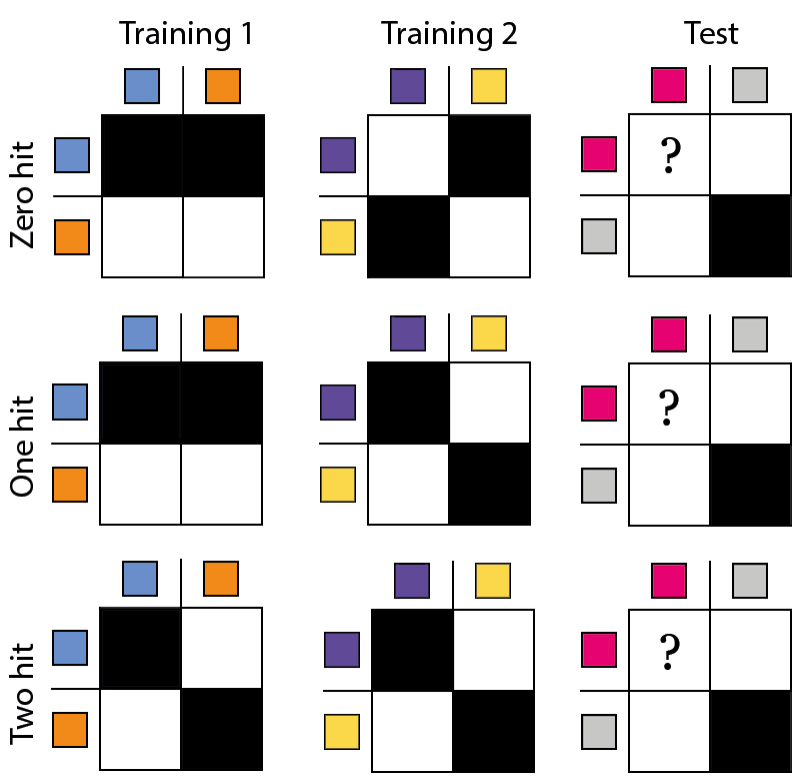}
    \caption{The three experimental conditions. The two leftmost columns give the two training games, and the rightmost column gives the testing game, which is the same for all conditions. The diagram for each game is a matrix, and gives its class-level relational structure. The entries of the matrix represent whether squares from different color classes interact (see Figure \ref{fig:blockmodel}). For the test games, the interaction indicated by a question mark, between elements of the first class, has not been observed at the first phase of questioning. The top row is the ``zero hit'' condition, and does not feature the testing game structure in any of the training games. The second row is the ``one hit'' condition, and features the test game structure once. The bottom row is the ``two hit'' condition, and features it both times.}
    \label{fig:two_conditions}
\end{figure}

Participants play two {training} games and then one test game. The underlying relational structures of the training and testing games in different conditions are shown in Figure \ref{fig:two_conditions}. In the ``zero hit'' condition, neither training game shares the underlying interactional structure of the test game. In the ``one hit'' and ``two hit'' conditions, one and two of the training games, respectively, have the same underlying structure as the testing game. In the \textit{training} games, participants observe all possible interactions between color classes and are visually presented the underlying class structure in a separate ``instructions'' diagram. The questions serve to guarantee they have acquired the full class-level structure of the game. In the first round of the \textit{test} game, participants have not observed any interactions between elements of one of the color classes; nor are they given an instructions diagram. This allows us to elicit their uncertainty over observed and unobserved class-level relations.

\subsection{Modeling details}
We connect our general modeling framework to the experimental tasks as follows. First, we model each interaction between pairs of blocks as a potentially biased coin flip---a Bernoulli random variable---that depends on the class of the two blocks in question. This applies to both the observed and the unobserved interactions, and we consider interactions independent given the model parameters and factor them accordingly:
\begin{align}
    r_{i, j}|\eta, {\bf z} \sim & \text{Bernoulli} (\eta_{{\bf z}_i, {\bf z}_j})\\
    p(\mathcal{R}|\eta, {\bf z}) = & \prod_{r\in\mathcal{R}} p(r|\eta, {\bf z}) \label{eqn:predictions}
\end{align}

The prior distributions over the class-assignment and class-interaction parameters, $\eta$ and ${\bf z}$, depend on the model. Recall that any particular model encodes abstract information about a previously encountered game in a class-level frequency vector, $f$, and class-level count, $u$, and non-count, $\bar{u}$ matrices for each relation. These data generate the model's parameters as follows:
\begin{align}
    {\bf \eta}_{A, B}|u, \bar{u}, \alpha \sim & \text{ Beta}(u_{A, B} + \alpha, \bar{u}_{A, B} + \alpha) \label{eqn:beta_analogy}\\
    {\bf z}|\zeta \sim & \text{ Multinomial }({\bf \zeta})\\
    \zeta| f, \lambda \sim & \text{ Dirichlet }(f + \lambda);
\end{align}
where $\alpha$ and $\lambda$ are hyperparameters used to smooth the prior distributions. $\zeta$ encodes a class-level frequency vector, which defines a prior for the class assignments.

These parameters define the posterior predictive distribution for a particular model as follows:
\begin{align}
& p(r|\mathcal{R}, M) = \sum_{{\bf z}'}\int_{\eta', 
\zeta'} p(r|\eta', 
\zeta', {\bf z}')\cdot p(\eta', 
\zeta', {\bf z}'|\mathcal{R}, M) \, d \eta' \, 
d\zeta'\\
& p(\eta, 
\zeta, {\bf z}|\mathcal{R}, M) = \frac{p(\mathcal{R}| \eta, 
\zeta, {\bf z}) \cdot p(\eta, 
\zeta, {\bf z}|M)}{p(\mathcal{R}|M)}\\
& p(\mathcal{R}| \eta, 
\zeta, {\bf z}) = p(\mathcal{R}| \eta, {\bf z})\\
& p(\eta, 
\zeta, {\bf z}|M) = p(\eta|u, \bar{u}, \alpha) \cdot p( 
{\bf z}|\zeta)\cdot p(\zeta|f, \lambda)
\end{align}


The last three lines show how the posterior over model parameters is calculated, and how the model's variables, $f, u, \bar{u}$, and hyperparameters, $\alpha$ and $\lambda$, factor in. We use conjugacies between the likelihood and prior distributions to collapse the model posterior:
\begin{align}
    p({\bf z}|\mathcal{R}, M) = \int_{\eta', \zeta'} p(\eta', 
\zeta', {\bf z}|\mathcal{R}, M) \, d \eta' \, d\zeta'
\end{align}

This allows us to then enumerate and sum over assignment vectors for each model to derive the final prediction, $p(r|\mathcal{R}, M)$. When we need to make predictions about an interaction we use the posterior mean over $\eta$ to provide a point estimate from this collapsed system \citep{kemp2010probabilistic}. 

When an individual has encountered more than one training environment, $\mathcal{M}\equiv \{ M^{(k)}\}_{k=1}^K$, these are each encoded as above. The posterior predictive distribution is now split over analogies as follows:
\begin{equation}
    p(r|\mathcal{R}) = \sum_{M'\in \mathcal{M}} p(r|\mathcal{R}, M')\cdot p(M'|\mathcal{R}).
\end{equation}

\noindent The model posterior---or, its mixture weight---is proportional to its evidence given the observed data multiplied by its prior weight:
\begin{equation}
    p(M|\mathcal{R}) \propto p(\mathcal{R}|M) \cdot p(M).
\end{equation}

For each experimental condition, we begin with two kernels representing the two structures that are consistent with the interactions observed in the testing game, phase 1. These are given virtual prior counts of 0.5, which models a prior preference for an interaction probability of 0.5 on the same-new question; we confirmed this figure during piloting (data not shown). We then append two additional kernels representing the structure of the training games a learner has encountered. These are given virtual prior counts of 1.  

The IRM is more fully presented elsewhere \citep{kemp2010probabilistic}. The main difference is in the prior distributions over theory components for a given model:
\begin{align}
    {\bf \eta}_{A, B}|u, \bar{u}, \alpha \sim & \text{ Beta}(\alpha, \alpha)\\
    {\bf z}|\gamma \sim & \text{ Chinese Restaurant Process }(\gamma)
\end{align}

\noindent The Chinese Restaurant Processes (CRP) is an exchangeable stochastic process that allows the assignment of elements to a potentially infinite number of classes \citep{aldous1985exchangeability}. In practice, it defines a probability for each possible \textit{partition}, or, clustering, of elements in the new system, favoring clusterings into fewer groups or classes. Given these assignments, no prior information is brought into making the class-level predictions (compare with the addition of the $u$ and $\bar{u}$ parameters in Equation \ref{eqn:beta_analogy}). Given a theory, probabilities are then assigned to interactions in the same way (see Equation \ref{eqn:predictions}). The functional consequences of these changes is that the IRM can compute probabilities for all interactions under all possible number of classes; but it does so na{\"i}vely, without including \textit{a priori} information on which specific class-level structures have been experienced by a learner before.

We extend our basic nonparametric model by including the IRM as one of our model kernels, allocating some of the prior probability mass, $\psi$, to it:
\begin{equation}
    p(M) = \begin{cases}
			\frac{1}{N+\psi}, & \text{each existing kernel}\\
            \frac{\alpha}{N+\psi} & \text{new kernel (IRM)}
		 \end{cases}
\end{equation}

In this study, we model inference using two sets of interactions, characterized by the relations ``interacts with'' and ``same color as''. These are provided in two sets of $u$ and $\bar{u}$ matrices, and are softened by separate $\alpha$ parameters. Models were relatively insensitive to the choice of hyperparameters $\alpha$, $\lambda$, and $\gamma$, and these were set at $0.01$, $0.01$, and $1$, respectively. The hyperparameter that controls the strength of the IRM kernel in the combined model, $\psi$, was fit using a grid search over the average relative entropy between the human empirical probabilities and the condition learners' estimated probabilities for each interaction (our estimated value was 600; see Results).

\begin{figure}[h!]
    \centering 
    \includegraphics[width=0.65\textwidth]{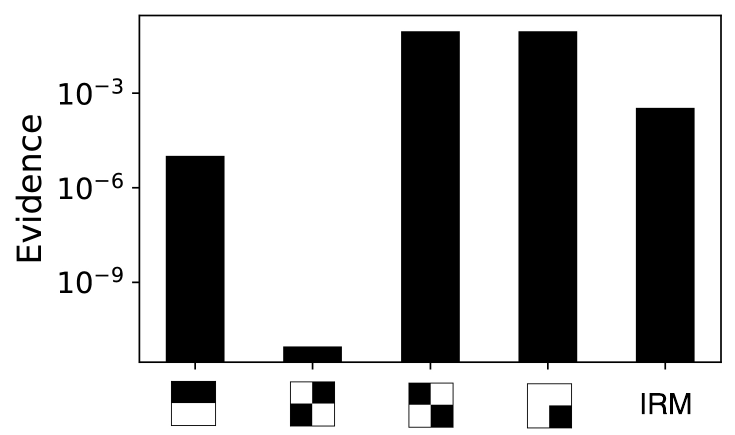}
    \caption{The model evidence for each model kernel given by the interactions and same-color relations observed by round one of the test game. Kernels are derived from the relational structure of the training games (bars 1-3), the kernels that are used to provide a baseline prior (bars 3-4), and the IRM.}
    \label{fig:model_evidence}
\end{figure}

We plot the evidence provided by the observed interactions, $p(\mathcal{R}|M)$, in round one of the test game for each model kernel in Figure \ref{fig:model_evidence}. The two generative structures that are inconsistent with some or all of the observed interactions receive low evidence (Figure \ref{fig:model_evidence}, first and second bar). The kernels representing the two consistent game structures receive high evidence (Figure \ref{fig:model_evidence}, third and fourth bar). The IRM also receives high evidence, as it includes the two consistent structures. However, it is more complex, and models a wider (exhaustive) distribution over structures, including inconsistent ones. This leads to a lower model evidence, which can also be thought of as a penalty for unnecessary complexity. 

\begin{figure}[h!]
    \centering 
    \includegraphics[width=\textwidth]{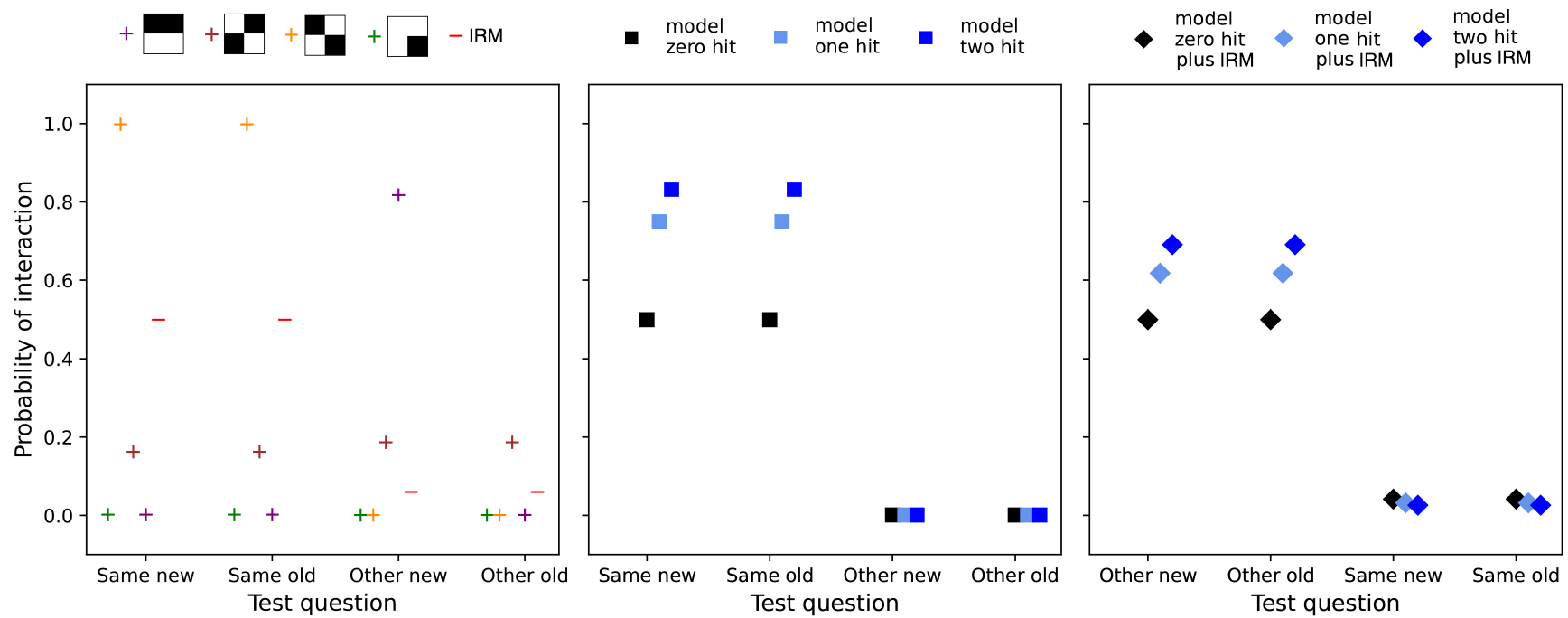}
    \caption{Model predictions. Left panel: the predictions generated from the basic structures of each training game, as well as the IRM fit on the interactions from those games, after hypothesis averaging. Middle panel: the predictions of our nonparametric model for each condition, after model averaging. Right panel: the predictions for each condition for the extended model, in which the IRM is included as a kernel and given a separate weight.}
    \label{fig:model_preds}
\end{figure}

We plot the model predictions in Figure \ref{fig:model_preds}. In the left panel, the predictions are generated according to the relational structures of each training or baseline game. In the middle panel these predictions are combined using the model evidence (see Figure \ref{fig:model_evidence}), and in the right panel the IRM is added as an additional kernel using the stochastic process prior described above. 

The key prediction made by our computational framework is that participants that have been exposed to the test game's relational structure during training will achieve higher scores. This is because the model kernel expressing the relational structure of the test game will receive the highest evidence, because it perfectly explains the test game's observed interactions (Figure \ref{fig:model_evidence}). Consequently, the predictions it makes as to the unobserved interaction (i.e., that the newly added square is very likely to interact with the existing square of the same color) will be weighted more highly during model averaging (Figure \ref{fig:model_preds}). 

On the other hand, if a participant has not encountered the test game's relational structure during training, the framework predicts they will be uncertain as to whether the interaction takes place (Figure \ref{fig:model_preds}). This is because the model kernels they have abstracted from training have low evidence given the test game's observed interactions (Figure \ref{fig:model_evidence}), and so inference is mainly controlled by the prior. 

According to this inferential framework, we expect participants to perform better if they have had prior exposure to a relational structure, recapitulating the finding of \citeauthor{gick1980analogical} (\citeyear{gick1980analogical}) in our novel visual paradigm. We also expect this effect to increase in a nonparametric manner, such that an increasing number of exposures to a particular structure increases the biases participants have towards expecting it again in the future. In the following section we use our visual blockworld paradigm to test these predictions, and examine the effect of different numbers of exposures to the relational structure of the test game.

\section{Methods}
\subsection{Participants}
We tested $200$ participants per condition, for a total of $600$. Participants for all experiments were recruited from Prolific, an online crowdsourcing platform that aims to ensure high-quality work (\url{https://www.prolific.co/}). Participants were excluded if English was not their first language, if they were not using a desktop computer, if their approval rate was below 95\%, and if they had taken part in any related studies by the authors. For each experiment there is a time-out period that automatically excludes participants, set by Prolific according to the expected median completion time. The number of participants per group was estimated using a power analysis from pilot studies, aimed at reproducing an effect size of $t\,=\,2$ under significance level of $97.5\%$ and a power of $80\%$.

Participants were paid bonuses in the rate phases of both training and testing games according to how well they scored. For each question of a rate phase, these bonuses depend on whether the block in question would light up or not, and scale linearly with a participant's response. If the block in the interaction posed in the question should light up, an answer of 0 (``definitely not'') would result in a bonus of $-1$\textcent, an answer of 5 would result in a bonus of $0$\textcent, and an answer of 10 (``definitely'') would result in a bonus of $1$\textcent. These bonuses were reversed if the interaction would not happen. For each of the training games, participants kept the final, highest bonus regardless of how many times they had to repeat. Participants were told that their bonuses in the test game will be multiplied by four, as it is harder to complete without instructions. Participants were paid a flat rate of $5$ dollars, and could earn bonuses up to $0.48$ dollars, for a total possible pay of $5.48$ dollars. The median completion time was $22.5$ minutes. 

This experiment was pre-registered at \url{www.aspredicted.org} under the title ``Analogy blockworlds two-hit'', number $122929$ (\url{https://aspredicted.org/cg4az.pdf}). 

\subsection{Procedure}
In all conditions, participants play two training games and one testing game. In the \textit{training} games, there are six square blocks, split equally into two groups of different colors. Participants are given a set of ``instructions'' for the game, which comprise a separate bounding box containing colored squares without letters, placed above the main game environment. These squares may be connected by arrows, which indicate whether elements of a particular color light up elements of their own and/or another color. An example manipulate phase from a training game is shown in Figure \ref{fig:demo_training}. In the rate phases, a block is selected from each color group at random. Participants are then asked about whether each block will light up and be lit up by a block from the same color group, and one from the alternate color group. They need to score more than 80\% of the total points to pass; otherwise they have to repeat the game until they do. This ensures a good knowledge of all possible interactions within the training game structure. When participants complete a training game they are told their score and bonus, and shown a second progress bar that indicated how many games they have completed relative to the total number.

In the \textit{test} game, participants are told that the game's instructions have been lost, and, consequently, it will be introduced more slowly. There are two rounds, each beginning with a manipulate phase without an instruction board. The first round begins with three blocks on screen, two from one color group and a third in its own color group. After an untimed exploratory phase, three new blocks are added such that each color group consists of three blocks. Before participants can move the new blocks, they are asked questions about how one of the new blocks will interact with the existing blocks (``test phase 1''; see Figure \ref{fig:demo_testing_rate}). Specifically, one of the new blocks from the color group previously containing one block was used as a probe object. Participants are asked, in random order, 1) whether the probe block would be lit up by the existing block from the same color group (``same new''); 2) whether it would light that block up (``same old''); 3) whether it would be lit up by an existing block from the other color group (``other new''); 4) whether it would light up the block from the other color class (``other old''). Participants then enter the manipulate phase of round 2, in which all six blocks from round 1 are present, before three new blocks are added and the same questions asked. After both phases they are asked to type how they think the blocks work. In the manipulate phase of round 2, all class-level interactions have been fully observed.

Participants' understanding of the experiment is tested in a quiz after the introductory instructions. The instruction quiz comprises the following multiple choice questions: how many games there will be (``3''); what participants will have to do in the game (``Move blocks around and then answer questions about their interactions''); how much bonus they would receive if two elements interact and they answered 0 (``$-1$\textcent''); 5 (``$0$\textcent''); or 10 (``$1$\textcent''); whether the relationship between the letters on blocks determine their interactions (``No''); whether all the games will have instructions (``No''); how much their bonus will change by if the games have lost their instructions (``Multiplied by four''); and, whether the blocks in the last game will work like the blocks in the previous games (``Maybe''). Participants must repeat this test until they score 100\%, reviewing which answers they answered incorrectly and the full set of instructions before every repeat.

Our main statistics of interest are participants' scores on each rating phase of the test game. For each phase, participants answer four questions about the new probe object's interactions with existing objects using the slider. If the probe object is called $P$, and the existing objects from either class $C1$ and $C2$, respectively, participants would answer: ``Consider what happens when $C \in \{C1, C2\}$ and P touch. How likely is $Z\in\{C, P\}$ to interact?'' For each of these questions, they could score a minimum of $-4$, and a maximum of $4$ points, giving a total minimum of $-16$ and a total maximum of $16$ points per rate phase. We measure the mean score in each rate phase per experimental condition, and compare these scores using an ANOVA $f$-test. We use independent pairwise ANOVA $t$-tests to measure differences between each of the means.

Participants indicate their response using a slider, with buttons labeled with increasing integers from 0 (``definitely not interact'') to 10 (``definitely interact''). We convert this to a prediction about the probability of different objects interacting by dividing all response values by 10. Then we calculate the expectation of the responses for each condition and question as follows:
\begin{align}
    p(\text{Object lights up})_{Human} &\equiv \mathbb{E}[v] \\
    & = \sum_{i=0}^{10} v_i \cdot p_i,
\end{align}
where $v_i$ is the new value of the slider button, a number between 0 and 1, and $p_i$ is the proportion of participants in that question and condition that gave that response as an answer. This allows us to compare human and model predictions using a single interaction probability. For our error bars, we begin by using the generic formula to calculate the standard deviation of responses given a particular condition and question: 
\begin{align}
    \sigma(\text{Object lights up})_{Human} \equiv \sqrt{\sum_{i=0}^{10} (v_i-\mathbb{E}[v])^2 \cdot p_i},
\end{align}
We then divide this figure by the square root of the number of judgments (${N=200})$ to obtain the standard error of the mean.

There are grounds for questioning any conclusions about aggregate or mean ordinal responses made using metric analyses \citep{liddell2018analyzing}. When we use response histograms to fit ordinal probit models and compare means in the resultant latent metric space all of our findings remain significant (data not shown).

\subsection{Stimuli}
Each blockworld comprises a $700\times700\text{px}$ bounding box containing $50\times50\text{px}$ colored blocks. When the blocks have a soft blue outline, they can be moved by clicking and dragging on them (manipulate phase). In the rate phases, they do not have a shadow, and are not movable. Between rounds one and two of the test game, three new blocks are added. When they are introduced, the new blocks are given a red outline. In this study, the blocks are always square, of the same size, named by letters, and turn red if they belong to interacting classes in a deterministic manner. Given these specifications (two classes of deterministically interacting color-based classes), there are 16 possible blockworlds. 

The color of different classes and the letters used to name blocks are randomized between trials and across participants. For each participant in all experiments, each game used different opponent colors for the blocks in different classes. These colors were randomized between an individual participant's training and test game, and between participants. The letters used to label different blocks were also shuffled between games and participants.

The experimental games are written in \texttt{javascript}, using a custom package for the \texttt{jsPsych} module \citep{de2015jspsych}. The experiment code itself is hosted on \url{www.heroku.com}, which \texttt{Prolific} routes participants to.

\subsection{Results}

\begin{figure}[h!]
    \centering 
    \includegraphics[width=0.6\linewidth]{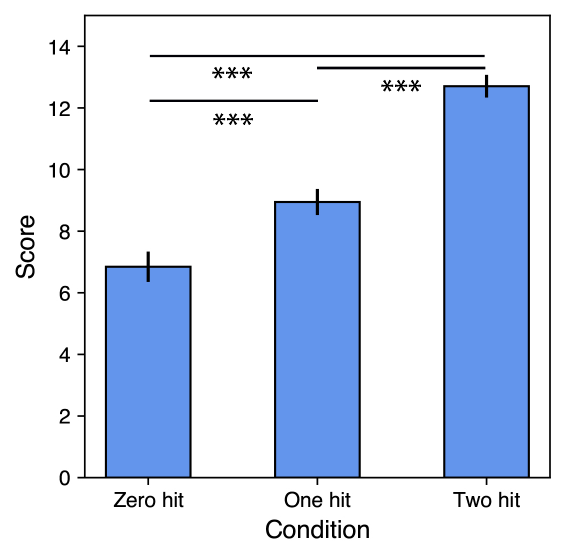}
    \caption{Average scores in the first testing round of the test game, split by condition. Participants scores in the one-hit are higher than the zero-hit condition, and are higher still in the two-hit condition. The error bars depict +/- 1 standard error of the mean. *** $\equiv \, p\,<\,0.001$.}
    \label{fig:two_scores}
\end{figure}

We present the mean score for each condition in round 1 of the test game in Figure \ref{fig:two_scores}. We observe that having experience with one training game with the same structure as the test game results in higher scores in the test game, and having experience with two training games strengthens this effect. An ANOVA across all conditions was highly significant ($f\,=\,47; p\,=\,1.1\text{e}-16$), as were all planned pairwise comparisons (one- \textit{versus} zero-hit condition, $t\,=\,3.46; p\,=\,0.0002$; two- \textit{versus} one-hit condition, $t\,=\,6.16; p\,=\,6.7\text{e}-10$). 

\begin{figure}[h!]
    \centering \includegraphics[width=0.85\linewidth]{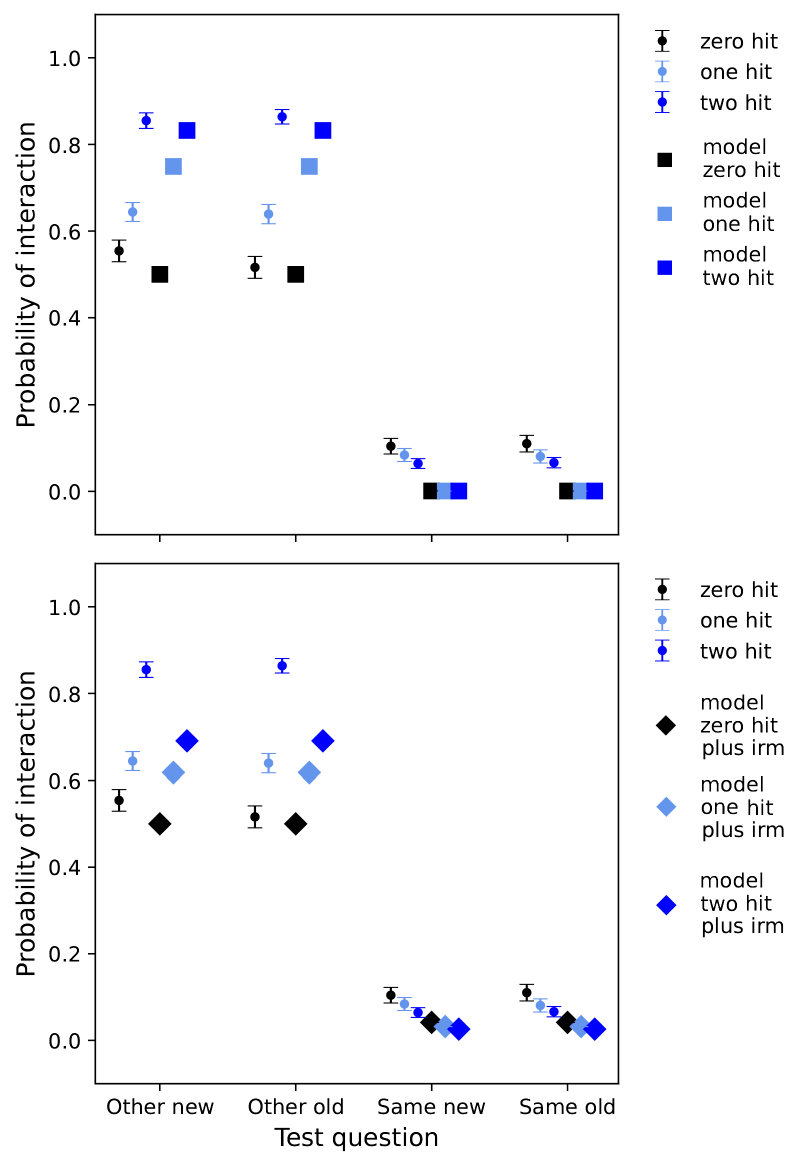}
    \caption{Model results. The top panel presents the nonparametric model predictions for all questions given the data observed at test phase 1, along with the aggregate human judgments for all conditions. The bottom panel presents the combined nonparametric model, which includes the IRM as one of its kernels. The error bars depict +/- 1 standard error of the mean.}
    \label{fig:two_model}
\end{figure}

We plot these results alongside predictions from our nonparametric modeling strategy in Figure \ref{fig:two_model} (top panel). For the same-class questions, the model predictions for the zero-hit and two-hit conditions match the human data well. For the one-hit condition, the models overestimate the probability of an interaction. For the other-class questions, our computational framework predicts an interaction probability of 0 for all conditions. However, on average participants allocate some probability to the blocks interacting across all conditions. One way of explaining this discrepancy is that participants were not entirely sure that color determines class, and so allocated some probability to the situation where the probe object may have belonged to a different class than the existing object of the same color. Interestingly, this is highest in the zero-hit condition, followed by the one-hit, followed by the two-hit, implying that participants' uncertainty about the null interaction was higher where a congruent model had not been encountered during training. This echoes a second important result in \citeauthor{gick1980analogical} (\citeyear{gick1980analogical}), in which participants exposed to a less consistent analogy during training generate a wider range of potential solutions to a verbal test problem. 

The IRM provides a natural model for this situation, as it gives a distribution over all possible theories that is flexible enough to capture such uncertainty. When we add the IRM as a kernel and examine the updated model predictions in Figure \ref{fig:two_model}, bottom row, we see an improved fit on almost all conditions and questions. The notable exceptions are the ``same'' questions for the two-hit condition, where the framework predicts a lower interaction probability than humans. It is possible that biases for higher-level patterns across sequences of training environments are influencing participants' predictions. That is, since the structure of the two training games has been identical and the observed test interactions are congruent with that pattern, participants may believe the game structure will always be the same, and only the colors differ, a consideration not currently included in our model, Although the quiz on the instructions emphasize that this is not necessarily the case, when we embed the two isomorphic training games into a three-game sequence with an incongruent game, the average response on these questions does decrease. By contrast, embedding the zero and one hit conditions in a three-game sequence does not affect the average prediction (data not shown). 

The strength of the IRM kernel in our extended model is fit with a mixing parameter, $\psi$. We found the best-fitting value for this parameter was $600$ (see Methods), which can be thought of as a prior weight. The prior weight of the model kernels encountered during training was, on the other hand, roughly $1$. In this experiment, the large value of $\psi$ was necessary to overcome the discrepancy between the model evidence for the IRM kernel and the model kernel abstracted from the structure of the test game (see Figure \ref{fig:model_evidence}). One way to explain this discrepancy is that the IRM is \textit{too} flexible, and does not focus its probability density on the (smaller) set of alternative theories people are considering. A more parsimonious model structure might explain participants' extra uncertainty relative to the simple structure-based model kernels. One suggestion for this would be a distribution that holds the relational structure between the three existing objects fixed, but allows the new object to also be allocated to its own class. In general, high values of parameters like $\psi$, which work to leave room for new theories or structures, could be used to suggest where new model kernels could be designed in order to better reflect the nature of the remaining uncertainty in participants' judgments.

In the second test round, there is no difference between the three conditions: participants are certain of the ground-truth test pattern, in which blocks only interact with blocks from their own color group. By this stage participants have observed every pair of color interactions, and are happy to generalize the relational pattern between classes to a new block from one of the existing color groups without observing it interact at all. This means they have abstracted the underlying relational structure, as well as the fact that color has perfectly predicted class in the interactions of this and previous games.


\section{Discussion}
In this paper we present a computational-level framework for relational generalization that takes into account both observed data from a new environment  and knowledge abstracted from environments a learner has encountered in the past. Our main contribution is to show how to pool information from a number of past environments in order to make better generalizations. Each past environment instantiates a kernel that defines a distribution over theories about the current environment, and the predictions from kernels are weighted by their relevance. In this way, generalization from both present and past experience can be viewed as nonparametric density estimation over explanatory structures. A main advantage is that this framework can incorporate other strategies for density estimation as additional kernels, such as more flexible theory learners like the IRM, and naturally extends to a stochastic process that can grow the number of kernels indefinitely over time. It also provides an indication of when an environment should be considered disjoint---as opposed to a continuation---through the relative posterior weight on the kernel representing a new explanation. 

This choice of prior corresponds to using a stochastic process---the CRP \citep{aldous1985exchangeability}---to define a nonparametric distribution over theories that can grow in complexity over time. When initially exploring an environment, the models of past environments---or, analogies---such a learner has access to would provide a strong inductive bias as to the types of structure they expect. In particular, if any past model is relevant given the new data, it should influence the learner's initial predictions. However, as observations in the new environment accrue, a more specific theory, acquired through a more flexible process like the IRM, will receive increasing evidence and dominate predictions. This phenomenon has been call the ``analogical shift conjecture'' \citep{gentner1983structure}, and is, perhaps, a part of the larger ``novice-expert shift'' \citep{carey1986cognitive}. 

Over successive environments the number of kernels should increase, such that the nonparametric distribution better reflects the structure of environments in a learner's world. This, in turn, will decrease the probability of instantiating new kernels. If our framework is an accurate model of human learning, it would provide complementary evidence that aspects of cognitive decline are a ``myth'', as suggested by Ramscar and colleagues (\citeyear{ramscar2014myth}). That is, according to our model the predictions and computational strategies of young and old learners will look quite different, as old ones try to reduce things to familiar structures, and young ones learn new structures from scratch. If the set of familiar structures cover a domain well, it may make more sense to rely on them for inference---at least initially. 

To our knowledge, this is the first time such a framework has been used as a cognitive model of generalization and analogy. In terms of generalization, it extends existing accounts of na{\"i}ve generalization based on features \citep{shepard1987toward}, discrete sets and multiple \textit{examples} \citep{tenenbaum2001generalization}, and relations \citep{kemp2010probabilistic, kemp2008discovery} to account for learning from multiple \textit{environments}. Although we use relational theories in this paper, our probabilistic framework is flexible enough to allow richer theories that express arbitrary combinations of continuous and discrete features, relations, and higher-order relations. A full ecological account must also explain why ``theories,'' ``kernels,'' or ``abstractions'' of environments take the form that they do---what the required generalizations were in previous environments, and why this particular information was abstracted to support them. In this study, the generalizations required in the training game were relational, and each relied on one prominent class-level structure. A more life-like situation might include multiple \textit{interpretations} of an environment, each of which forms a basis or kernel for generalization and can be updated independently \citep{lakoff2008metaphors}. 

An as-yet unexplored advantage of our computational framework is that different models can bind to data using entirely different parameter sets. Consider the different knowledge required to ride, fix, or design a bicycle. In these extremes, it may be necessary to define different \textit{parameter sets} for different kernels---in the case of riding, physics engines \citep{ullman2017mind}; in the case of fixing, reinforcement learning \citep{tsividis2021human}; and in the case of designing, probabilistic programs \citep{lake2015human, ellis2020dreamcoder}.



In terms of analogy, we show that seminal experimental results from cognitive psychology extend to the setting of multiple environments \citep{gick1980analogical}. We are also able to give ecological justifications for several important computational-level assumptions of SMT \citep{gentner1983structure}. In considering assignments of \textit{elements} in the new environment to the \textit{classes} of an existing model, we relax the one-to-one constraint that is normally imposed on analogical mappings, allowing one-to-many and many-to-one assignments. Under our formalism, one-to-one assignments are only favored when it is warranted by the data. A second important aspect of SMT is ranking mappings in terms of their \textit{systematicity} (see Introduction). In our framework, systematicity is addressed at two levels: the relevance---or, posterior probability---of element assignments during hypothesis averaging, {and} the relevance of entire models of past environments during model averaging. This allows us to consider how generalizations from multiple environments can be combined together---a phenomenon not explicitly addressed in most models of analogy between pairs of concepts or domains. The greater the the depth of match between the generative models serving as abstractions of different environments, the more accurate---the more \textit{systematic}---their predictions about the other should be \citep{kemp2005generative}. Models with higher evidence are by definition more systematic matches; but, as we have seen above, the prior allows factors like normality and epistemic uncertainty to play a role.

In terms of statistics, model comparisons and combinations are the basis of data analysis in most scientific disciplines, and the norm for applied fields such as economics and finance. However, our analysis is unusual in that it explicitly addressed the problem of how to combine predictions from potentially very different model classes. This approach has been developed in more theoretical detail in three advanced settings related to statistical inference. The first examines shows how to aggregate predictions from convex combinations of exponential-family distributions when making inferences on graphs \citep{wainwright2002stochastic}. The second two are latent variable models and show how to aggregate predictions from disjoint clusters of latent variables \citep{hoffman2015structured, mansinghka2016crosscat}. In some sense, our framework is a generalization of these approaches, to include a much wider range of model families, each generated by a kernel. But the cost of this generality is a lack of tightly controlled mathematical properties that can be used to provide bounds and guarantee behavior. An interesting theoretical extension would be to examine cases where computing or approximating several smaller log partition functions is preferable to one larger one---in terms of both accuracy and computational complexity. Here, too, the idea of different kernels using different parameter sets is attractive---partition functions are a function of the number of datapoints \textit{and} the model parameters. 

In spirit this study is closest to recent work in machine learning. In particular, the data-driven extraction of analogical mappings from vector spaces \citep{lu2022probabilistic, webb2023zero} and the abstraction and use of structured world models in solving complex reinforcement learning (RL) tasks \citep{hafner2020mastering}. Indeed, in related work exploring theory-based reinforcement learning, \citeauthor{pouncy2022inductive} (\citeyear{pouncy2022inductive}) employ a similar formalism in order to efficiently approximate complex task structures using a technique developed in \citeauthor{saeedi2017variational} (\citeyear{saeedi2017variational}). We are proposing a combination of these approaches: explicitly abstracting world models from multiple training environments, and then generalizing these world models to novel situations in service of inference. A major next step will be to collect human behavior over trajectories of more complex RL tasks, and extend our framework to try to explain their {actions}.

\bibliography{thesis}

@article{lu2022probabilistic,
  title={Probabilistic analogical mapping with semantic relation networks.},
  author={Lu, Hongjing and Ichien, Nicholas and Holyoak, Keith J},
  journal={Psychological Review},
  volume={129},
  number={5},
  pages={1078},
  year={2022},
  publisher={American Psychological Association}
}

@incollection{aldous1985exchangeability,
	title        = {Exchangeability and related topics},
	author       = {Aldous, David J},
	year         = 1985,
	booktitle    = {{\'E}cole d'{\'E}t{\'e} de Probabilit{\'e}s de {S}aint-{F}lour {XIII}—1983},
	publisher    = {Springer},
	pages        = {1--198}
}

@article{carey1986cognitive,
	title        = {Cognitive science and science education.},
	author       = {Carey, Susan},
	year         = 1986,
	journal      = {American {P}sychologist},
	publisher    = {American Psychological Association},
	volume       = 41,
	number       = 10,
	pages        = {1123--1130}
}

@article{de2015jspsych,
	title        = {js{P}sych: A {J}avaScript library for creating behavioral experiments in a {W}eb browser},
	author       = {De Leeuw, Joshua R},
	year         = 2015,
	journal      = {Behavior {R}esearch {M}ethods},
	publisher    = {Springer},
	volume       = 47,
	pages        = {1--12}
}

@article{ellis2020dreamcoder,
	title        = {Dreamcoder: Growing generalizable, interpretable knowledge with wake-sleep {B}ayesian program learning},
	author       = {Ellis, Kevin and Wong, Catherine and Nye, Maxwell and Sable-Meyer, Mathias and Cary, Luc and Morales, Lucas and Hewitt, Luke and Solar-Lezama, Armando and Tenenbaum, Joshua B},
	year         = 2020,
	journal      = {arXiv preprint arXiv:2006.08381}
}

@article{gentner1983structure,
	title        = {Structure-mapping: A theoretical framework for analogy},
	author       = {Gentner, Dedre},
	year         = 1983,
	journal      = {Cognitive {S}cience},
	publisher    = {Elsevier},
	volume       = 7,
	number       = 2,
	pages        = {155--170}
}

@article{gick1980analogical,
	title        = {Analogical problem solving},
	author       = {Gick, Mary L and Holyoak, Keith J},
	year         = 1980,
	journal      = {Cognitive {P}sychology},
	publisher    = {Elsevier},
	volume       = 12,
	number       = 3,
	pages        = {306--355}
}

@book{gopnik1997words,
	title        = {Words, thoughts, and theories},
	author       = {Gopnik, Alison and Meltzoff, Andrew N and Bryant, Peter},
	year         = 1997,
	publisher    = {MIT Press Cambridge, MA},
	volume       = 1
}

@article{hafner2020mastering,
	title        = {Mastering atari with discrete world models},
	author       = {Hafner, Danijar and Lillicrap, Timothy and Norouzi, Mohammad and Ba, Jimmy},
	year         = 2020,
	journal      = {arXiv preprint arXiv:2010.02193}
}

@inproceedings{hoffman2015structured,
	title        = {Structured stochastic variational inference},
	author       = {Hoffman, Matthew D and Blei, David M},
	year         = 2015,
	booktitle    = {Artificial {I}ntelligence and {S}tatistics},
	pages        = {361--369}
}

@book{hofstadter2008metamagical,
	title        = {Metamagical themas: Questing for the essence of mind and pattern},
	author       = {Hofstadter, Douglas R},
	year         = 2008,
	publisher    = {Hachette UK}
}

@article{holland1983stochastic,
	title        = {Stochastic blockmodels: First steps},
	author       = {Holland, Paul W and Laskey, Kathryn Blackmond and Leinhardt, Samuel},
	year         = 1983,
	journal      = {Social {N}etworks},
	publisher    = {Elsevier},
	volume       = 5,
	number       = 2,
	pages        = {109--137}
}

@article{kass1995bayes,
	title        = {Bayes factors},
	author       = {Kass, Robert E and Raftery, Adrian E},
	year         = 1995,
	journal      = {Journal of the {A}merican {S}tatistical {A}ssociation},
	publisher    = {Taylor \& Francis},
	volume       = 90,
	number       = 430,
	pages        = {773--795}
}

@article{kemp2010probabilistic,
	title        = {A probabilistic model of theory formation},
	author       = {Kemp, Charles and Tenenbaum, Joshua B and Niyogi, Sourabh and Griffiths, Thomas L},
	year         = 2010,
	journal      = {Cognition},
	publisher    = {Elsevier},
	volume       = 114,
	number       = 2,
	pages        = {165--196}
}

@article{kemp2008discovery,
	title        = {The discovery of structural form},
	author       = {Kemp, Charles and Tenenbaum, Joshua B},
	year         = 2008,
	journal      = {Proceedings of the National Academy of Sciences},
	publisher    = {National Acad Sciences},
	volume       = 105,
	number       = 31,
	pages        = {10687--10692}
}

@article{lake2015human,
	title        = {Human-level concept learning through probabilistic program induction},
	author       = {Lake, Brenden M and Salakhutdinov, Ruslan and Tenenbaum, Joshua B},
	year         = 2015,
	journal      = {Science},
	publisher    = {American Association for the Advancement of Science},
	volume       = 350,
	number       = 6266,
	pages        = {1332--1338}
}

@book{lakoff2008metaphors,
	title        = {Metaphors we live by},
	author       = {Lakoff, George and Johnson, Mark},
	year         = 2008,
	publisher    = {University of Chicago Press}
}

@article{liddell2018analyzing,
	title        = {Analyzing ordinal data with metric models: What could possibly go wrong?},
	author       = {Liddell, Torrin M and Kruschke, John K},
	year         = 2018,
	journal      = {Journal of Experimental Social Psychology},
	publisher    = {Elsevier},
	volume       = 79,
	pages        = {328--348}
}

@article{mansinghka2016crosscat,
	title        = {Crosscat: A fully {B}ayesian nonparametric method for analyzing heterogeneous, high dimensional data},
	author       = {Mansinghka, Vikash and Shafto, Patrick and Jonas, Eric and Petschulat, Cap and Gasner, Max and Tenenbaum, Joshua B},
	year         = 2016,
	publisher    = {MIT Press}
}

@article{nosofsky1986attention,
	title        = {Attention, similarity, and the identification--categorization relationship.},
	author       = {Nosofsky, Robert M},
	year         = 1986,
	journal      = {Journal of {E}xperimental {P}sychology: General},
	publisher    = {American Psychological Association},
	volume       = 115,
	number       = 1,
	pages        = {39--61}
}

@article{pouncy2022inductive,
	title        = {Inductive biases in theory-based reinforcement learning},
	author       = {Pouncy, Thomas and Gershman, Samuel J},
	year         = 2022,
	journal      = {Cognitive Psychology},
	publisher    = {Elsevier},
	volume       = 138,
	pages        = 101509
}

@article{saeedi2017variational,
	title        = {Variational particle approximations},
	author       = {Saeedi, Ardavan and Kulkarni, Tejas D and Mansinghka, Vikash K and Gershman, Samuel J},
	year         = 2017,
	journal      = {The Journal of Machine Learning Research},
	publisher    = {JMLR. org},
	volume       = 18,
	number       = 1,
	pages        = {2328--2356}
}

@article{shepard1987toward,
	title        = {Toward a universal law of generalization for psychological science},
	author       = {Shepard, Roger N},
	year         = 1987,
	journal      = {Science},
	publisher    = {American Association for the Advancement of Science},
	volume       = 237,
	number       = 4820,
	pages        = {1317--1323}
}

@inproceedings{kemp2005generative,
  title={A generative theory of similarity},
  author={Kemp, Charles and Bernstein, Aaron and Tenenbaum, Joshua B},
  booktitle={Proceedings of the {A}nnual {M}eeting of the {C}ognitive {S}cience {S}ociety},
  pages={1132--1137},
  year={2005},
  organization={Citeseer}
}

@article{tenenbaum2006theory,
	title        = {Theory-based {B}ayesian models of inductive learning and reasoning},
	author       = {Tenenbaum, Joshua B and Griffiths, Thomas L and Kemp, Charles},
	year         = 2006,
	journal      = {Trends in Cognitive Sciences},
	publisher    = {Elsevier},
	volume       = 10,
	number       = 7,
	pages        = {309--318}
}

@article{tenenbaum2001generalization,
	title        = {Generalization, similarity, and {B}ayesian inference},
	author       = {Tenenbaum, Joshua B and Griffiths, Thomas L},
	year         = 2001,
	journal      = {Behavioral and Brain Sciences},
	publisher    = {Cambridge University Press},
	volume       = 24,
	number       = 4,
	pages        = {629--640}
}

@article{tsividis2021human,
	title        = {Human-level reinforcement learning through theory-based modeling, exploration, and planning},
	author       = {Tsividis, Pedro A and Loula, Joao and Burga, Jake and Foss, Nathan and Campero, Andres and Pouncy, Thomas and Gershman, Samuel J and Tenenbaum, Joshua B},
	year         = 2021,
	journal      = {arXiv preprint arXiv:2107.12544}
}

@article{ullman2017mind,
	title        = {Mind games: Game engines as an architecture for intuitive physics},
	author       = {Ullman, Tomer D and Spelke, Elizabeth and Battaglia, Peter and Tenenbaum, Joshua B},
	year         = 2017,
	journal      = {Trends in {C}ognitive {S}ciences},
	publisher    = {Elsevier},
	volume       = 21,
	number       = 9,
	pages        = {649--665}
}

@phdthesis{wainwright2002stochastic,
	title        = {Stochastic processes on graphs with cycles: geometric and variational approaches},
	author       = {Wainwright, Martin James},
	year         = 2002,
	chapter      = 7,
	school       = {Massachusetts Institute of Technology}
}

@article{webb2023zero,
  title={Zero-shot visual reasoning through probabilistic analogical mapping},
  author={Webb, Taylor and Fu, Shuhao and Bihl, Trevor and Holyoak, Keith J and Lu, Hongjing},
  journal={Nature Communications},
  volume={14},
  number={1},
  pages={5144},
  year={2023},
  publisher={Nature Publishing Group UK London}
}

@article{ramscar2014myth,
  title={The myth of cognitive decline: Non-linear dynamics of lifelong learning},
  author={Ramscar, Michael and Hendrix, Peter and Shaoul, Cyrus and Milin, Petar and Baayen, Harald},
  journal={Topics in {C}ognitive {S}cience},
  volume={6},
  number={1},
  pages={5--42},
  year={2014},
  publisher={Wiley Online Library}
}







\end{document}